\documentclass[10pt,twocolumn,letterpaper]{article}

\usepackage{cvpr}              

\definecolor{cvprblue}{rgb}{0.21,0.49,0.74}
\usepackage[pagebackref,breaklinks,colorlinks,allcolors=cvprblue]{hyperref}
\usepackage{multirow}

\def\paperID{5539} 
\def\confName{CVPR}
\def\confYear{2025}

\title{UMIT: Unifying Medical Imaging Tasks via Vision-Language Models}

\author{Haiyang Yu\thanks{Equal Contribution \newline \hspace*{0.35cm}\dag Corresponding Author}\ , Siyang Yi$^{*}$, Ke Niu, Minghan Zhuo, Bin Li$^{\dag}$\\
Shanghai Key Laboratory of Intelligent Information Processing\\ School of Computer Science, Fudan University\\
{\tt\small hyyu20@fudan.edu.cn, \{yisy22, kniu22, mhzhuo24\}@m.fudan.edu.cn}\\
{\tt\small libin@fudan.edu.cn}}

\usepackage{tcolorbox}
\usepackage{xcolor} 
\definecolor{whitesmoke}{HTML}{F5F5F5}

\begin{document}
\maketitle
\begin{abstract}

With the rapid advancement of deep learning, particularly in the field of medical image analysis, an increasing number of Vision-Language Models (VLMs) are being widely applied to solve complex health and biomedical challenges. However, existing research has primarily focused on specific tasks or single modalities, which limits their applicability and generalization across diverse medical scenarios. To address this challenge, we propose \textbf{UMIT}, a unified multi-modal, multi-task VLM designed specifically for medical imaging tasks. UMIT is able to solve various tasks, including visual question answering, disease detection, and medical report generation. In addition, it is applicable to multiple imaging modalities (\textit{e.g.}, X-ray, CT and PET), covering a wide range of applications from basic diagnostics to complex lesion analysis. Moreover, UMIT supports both English and Chinese, expanding its applicability globally and ensuring accessibility to healthcare services in different linguistic contexts. To enhance the model’s adaptability and task-handling capability, we design a unique two-stage training strategy and fine-tune UMIT with designed instruction templates. Through extensive empirical evaluation, UMIT outperforms previous methods in five tasks across multiple datasets. The performance of UMIT indicates that it can significantly enhance diagnostic accuracy and workflow efficiency, thus providing effective solutions for medical imaging applications. The code and adopted datasets are available at \href{https://github.com/dz-osamu/UMIT/tree/main}{https://github.com/dz-osamu/UMIT/tree/main}.
\end{abstract}    
\section{Introduction}
\label{sec:intro}

Traditional medical imaging tasks typically rely on specialized models, which can only perform excellently on individual tasks. However, they lack cross-task generalization, reducing flexibility and efficiency in practical applications. In recent years, Vision-Language Models (VLMs) have significantly enhanced the ability to understand images and generate text by combining visual and linguistic information, demonstrating exceptional performance across various multi-modal tasks~\cite{liu2024visual,alayrac2022flamingo,wang2024qwen2}. In the medical field, an increasing number of studies are introducing VLMs for medical image analysis, aiming at achieving more intelligent and efficient multi-task processing.

\begin{figure}[t]
    \centering
    \includegraphics[width=1\linewidth]{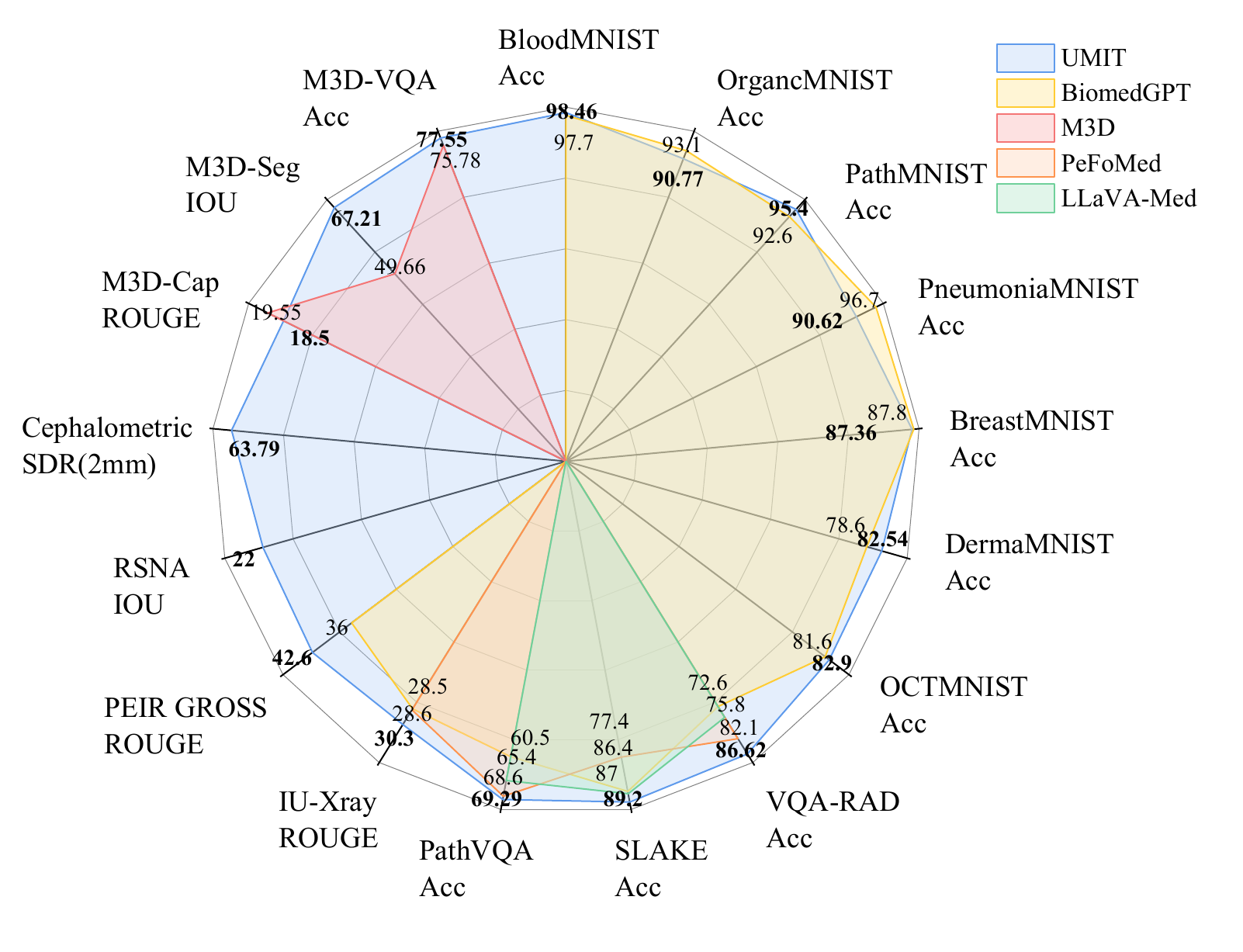}
\caption{Performance comparison with various models. UMIT utilizes all the datasets and demonstrates broad applicability across multiple tasks and modalities.}
\label{fig:intro}
\end{figure}

Existing Vision-Language Models (VLMs) for medical image analysis can be divided into two main categories: CLIP-based approach and LLM-based approach. For the CLIP-based approach, the training process primarily focuses on contrastive learning between images and text~\cite{radford2021learning}. Therefore, they have difficulties in handling complex medical tasks such as disease localization and report generation. For example, PLIP~\cite{huang2023visual} trains CLIP on a large set of pathology image-text pairs for zero-shot classification task. BiomedCLIP~\cite{zhang2023biomedclip} introduces a CLIP-based foundation model adaptable to various medical imaging modalities, applying it to classification and Visual Question Answering (VQA) tasks. On the other hand, the LLM-based approach typically uses the language model as decoder and formulates medical multi-modal tasks into target sequence generation. This approach not only has stronger contextual understanding and reasoning capabilities, but also can adapt to different medical tasks through fine-tuning. However, existing LLM-based methods still fall short in the breadth of applications. For example, LLaVA-Med~\cite{li2024llava} is the first model to introduce multi-stage, multi-modal instructions for visual-text data analysis in the biomedical field, but it only supports English and is limited to VQA tasks. RadFM~\cite{wu2023towards} is the first system trained on 2D and 3D medical datasets, which is mainly oriented to text generation tasks.

In this paper, we propose a versatile multi-modal, multi-task Vision-Language Model called \textbf{UMIT}. Specifically, UMIT is designed to handle a wide range of critical medical tasks, including medical VQA, report generation, image classification, disease detection, and landmark detection. These tasks cover most medical automating requirements, providing strong support for clinical decision-making, report generation, and diagnosis. Furthermore, the proposed UMIT is capable of processing both 2D and 3D medical images and supports bilingual inputs, significantly expanding its potential applications across various medical domains. As shown in Figure~\ref{fig:intro}, UMIT involves a broader range of datasets compared to other models, covering various tasks and medical imaging modalities. By leveraging its ability to process diverse modalities and perform multiple tasks, UMIT offers a comprehensive and adaptable solution that can enhance the efficiency and accuracy of medical image interpretation, diagnostic workflows, and report generation processes. To achieve this goal, we design a two-stage training strategy: feature alignment stage and instruction fine-tuning stage. Through experiments, UMIT outperforms previous SOTA methods across multiple datasets in five tasks. The main contributions of this paper can be summarized as follows:

\begin{itemize}
\item We propose a Vision-Language Model UMIT, which is capable of handling multi-modal and multi-task medical imaging applications.

\item We design a two-stage training strategy: feature alignment stage and instruction fine-tuning stage. The feature alignment stage aims at aligning visual and textual features while the fine-tuning stage employs various instructions to enable UMIT to share knowledge across multiple tasks.

\item We conduct experiments on multiple public benchmark datasets, including five tasks and eighteen datasets. The results demonstrate that UMIT outperforms the state-of-the-art methods on most datasets. 
\end{itemize}
\section{Related Work}
\label{sec:related work}

\subsection{Vision-Language Models}
Inspired by the success of Large Language Models (LLMs) like GPT~\cite{ouyang2022training}, various Vision-Language Models (VLMs) have been developed to extract information from both visual and textual modalities, overcoming the limitations of LLMs that primarily focus on textual data. Currently, the most widely used approaches for VLMs can be broadly categorized into two main types: CLIP-based approaches and LLM-based approaches.

CLIP (Contrastive Language-Image Pretraining)~\cite{radford2021learning}, pre-trained using a large amount of natural language paired with image data through contrastive learning, aligns the representations of images and text. Models developed from CLIP, such as PLIP~\cite{huang2023visual}, PMC-CLIP~\cite{lin2023pmc} and BiomedCLIP~\cite{zhang2023biomedclip} in the medical field, are therefore able to analyze pathology images, classify disease types, retrieve corresponding images for given textual descriptions and complete other tasks related to image-text matching. However, since CLIP-based models use contrastive learning for image-text alignment, they lack the ability to generate content, which makes it difficult for them to handle tasks such as position localization and report generation.

On the other hand, LLM-based VLMs, taking Qwen2-VL~\cite{wang2024qwen2} as an example, perform well in complex and diverse tasks. LLM-based VLMs generally extract features through a visual encoder and a cross-modal connector, and then passing them to the LLM for further processing. Based on this, Qwen2-VL introduces modules such as Native Dynamic Resolution mechanism and multi-modal Rotary Position Embedding, further enhancing its ability to process, percept and generate multi-modal information. Therefore, Qwen2-VL suits for various types of tasks including VQA, report generation and other specific tasks that required complex imaging processing ability. As for the utilizations of LLM-based VLMs in medical field, there have emerged several models to tackle certain challenges, and models that aim to accomplish more comprehensive tasks mostly take LLM-based approaches as their methods because of its strong information processing capacity.




\begin{figure*}[t]
    \centering
    \includegraphics[width=0.95\linewidth]{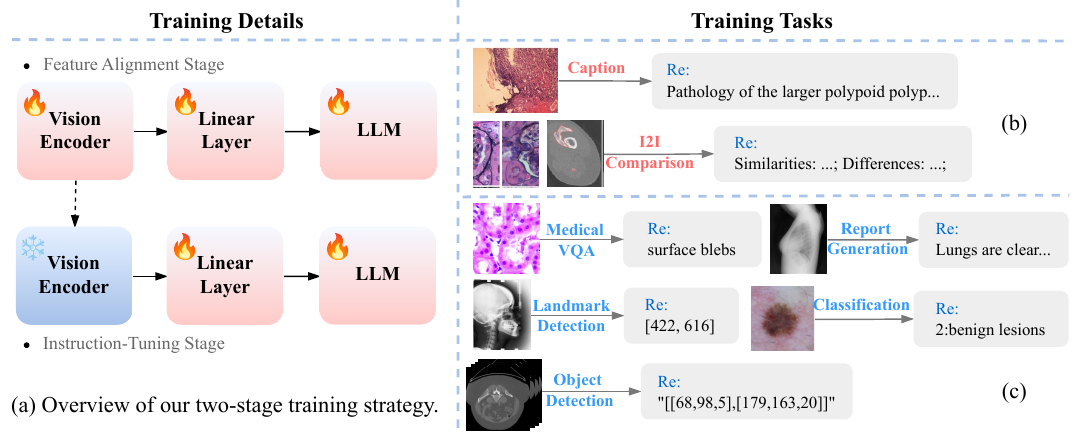}
\caption{The framework of the proposed UMIT. (a) Overview of our two-stage training strategy. (b) Training tasks in feature alignment stage. (c) Training tasks in instruction-tuning stage.}
\label{framework}
\end{figure*}

\subsection{Unified VLMs for Medical Image Analysis}
In the field of medical image analysis, the complex and diverse data is challenging for traditional machine learning models to handle. Medical analysis requires not only the interpretation of different types of features, including features extracted from images and text, but also the knowledge of professional medical field. At present, VLMs have shown the potential to achieve these requirements. 
It is important to note that real-world medical image analysis tasks contain a wide variety of tasks, and need multiple language supports for wider applications. 
Therefore, it is necessary to introduce models that are capable of handling multi-modal, multilingual and multi-task data, which we refer to as unified VLMs.

Recently, a number of unified VLMs, primarily LLM-based VLMs, that focused on meeting the multi-modal and multi-task requirements of medical image analysis have been developed.
However, these methods still have limitations, such as restricted task coverage, limited compatibility with medical imaging modalities, or incomplete language support.
Specifically, to match the multi-task requirements, 
PeFoMed~\cite{he2024pefomed} provides a solution for Med-VQA and medical report generation through a parameter-effective fine-tuning approach from general-domain transferred to medical image analysis. 
GSCo~\cite{he2024gsco} combines the context learning ability of unified models with the domain expertise of specialist models, and introduces two collaborative mechanisms, Mixture-of-Expert Diagnosis and Retrieval-Augmented Diagnosis, effectively enhancing the performance of VQA, report generation, and medical image diagnosis tasks.
By integrating cross-disciplinary data, BiomedGPT~\cite{zhang2024generalist} has achieved state-of-the-art performance metrics across multiple datasets for classification, VQA, and report generation tasks.
To accommodate various medical imaging modalities, such as 2D X-rays, ultrasounds, and 3D CTs, MRIs, etc.
RadFM~\cite{wu2023towards} proposes a generalist foundation fine-tuning model specifically designed for VQA and report generation tasks in radiology, which can be applied to both 2D and 3D images. 
Regarding multilingual models, Qilin-Med-VL~\cite{liu2023qilin} constructs the first large Chinese VLM for general healthcare tasks, using a translated version of English open source datasets provided by GPT-3.5 with quality control. HuatuoGPT-Vision~\cite{chen2024huatuogpt} constructed a high-quality multi-modal medical dataset based on PubMed\footnote{PubMed is a free search engine that primarily accesses the MEDLINE database, containing references and scientific papers on life sciences and biomedical topics.} and trained a VLM capable of bilingual medical VQA using this dataset.

Although these models have been successfully applied to various scenarios, their scalability across different medical domains and tasks remains limited. To address this, we propose UMIT, a state-of-the-art unified VLM specifically designed for medical image analysis, capable of efficiently handling diverse tasks across multiple modalities in both Chinese and English.



\section{Methodology}
\label{sec:method}
\subsection{Framework}


Since Qwen2-VL~\cite{wang2024qwen2} supports bilingual prompts and has robust image and video understanding capabilities, we simply employ the model architecture of Qwen2-VL. As shown in Figure~\ref{framework}, UMIT includes three modules: a vision encoder, a linear layer, and a language model as decoder. The linear layer acts as the connector to map the channel dimension between the vision encoder and the language model. We design a two-stage training strategy to optimize the whole model. In the first stage, we pre-train the model using a large number of medical image-text pairs. This stage helps establish a connection between the visual features of the medical images and the textual features containing specialized knowledge, achieving feature alignment. In the second stage, we design multiple instructions to fine-tune the model, enabling it establish connections between different tasks. The designed instructions can be adjusted according to the specific requirements of each task and modality, thereby enhancing its flexibility and adaptability in multi-task and multi-modal environments.

\subsection{Feature Alignment Stage}
In the feature alignment stage, the goal is to enable UMIT to learn how to align visual and textual information in a shared feature space. For example, in medical imaging tasks, an X-ray image may correspond to a medical report or diagnostic description associated with the image. We initialize UMIT with Qwen2-VL pre-trained weights, and train all parameters of the model to ensure it effectively learn the relationships between medical images and text. This approach not only enhances the model's generalization ability for downstream tasks but also improves the consistency of its understanding of fundamental medical concepts.

\begin{table*}[t]
\raggedright
    \centering
    \caption{Description of all datasets used by UMIT for training and testing during Instruction-Tuning phase}
    \label{tab:Fine-tuning}
    \scalebox{0.75}{
        \begin{tabular}{cp{3cm}p{12cm}c} %
            \toprule 
            Task & Dataset & Description & Train/valid/test split \\ 
            \midrule 
            \multirow{9}{*}{Medical VQA} & VQA-RAD &  A medical visual question-answering dataset consists of open-ended and close-ended question–answer pairs on 315 2D radiology images. &1,797 / 0 / 451 \\ 
            & SLAKE & An English-Chinese bilingual dataset consisting of 642 2D radiology images. & 9,835 / 2,099 / 2,094\\
            & PathVQA & A medical visual question-answering dataset consists of open-ended questions from 4,998 2D pathology images. & 19,654 / 6,279 / 6,719\\
            & M3D-VQA & A medical visual question-answering dataset adopts the multiple-choice format based on 3D images. & 479,664 / 14,068 / 13,791 \\
            & CM-VL-I & A Chinese medical visual question-answering dataset which contain a diverse collection of data. & 150,603 / 0 / 2,000 \\
            \midrule
            \multirow{13}{*}{Classification} & PathMNIST &  A histopathological image classification dataset consists of multi-class classification tasks on 107,180 image patches from colorectal cancer histology slides.  &89,996 / 10,004 / 7,180 \\ 
            & DermaMNIST &  A dermatology image classification dataset consists of 10,015 dermatoscopic images categorized into 7 disease types.  &7,007 / 1,003 / 2,005 \\ 
            & OCTMNIST &  An optical coherence tomography image classification dataset consists of 109,309 gray-scale images for retinal disease diagnosis, categorized into 4 classes.  &97,477 / 10,832 / 1000 \\ 
            & PneumoniaMNIST &  A pediatric chest X-ray image dataset consists of 5,856 gray-scale images for a binary classification task of pneumonia versus normal.  &4,708 / 524 / 624 \\ 
            & BreastMNIST &  A breast ultrasound image dataset consists of 780 images categorized into 3 classes, simplified into a binary classification task.  &546 / 78 / 156 \\ 
            & BloodMNIST &  A blood cell image dataset consists of 17,092 images categorized into 8 classes.  &11,959 / 1,712 / 3,421 \\ 
            & OrgancMNIST &  An organ classification dataset based on 3D CT images consists of 2D gray-scale slices from 11 body organs for multi-class classification task.  &12,975 / 2,392 / 8,216 \\ 
            \midrule
            \multirow{6}{*}{Report Generation} & IU-Xray & A medical dataset of 7,470 2D chest X-rays and corresponding diagnostic reports detailing the patient's history, symptoms, radiologic findings, and final diagnosis.  & 2,069 / 590 / 296 \\
            & PEIR GROSS &  A medical dataset consists of 7,442 2D image-caption pairs collected from the Gross sub-collection of the PEIR digital library, categorized into 21 different sub-categories.   & 6,697 / 0 / 745 \\ 
            & M3D-Cap &  A public medical dataset consisting of 3D CT medical images and corresponding expert-reviewed reports collected from publicly accessible medical websites  & 115,459 / 2,000 / 2,000 \\
            \midrule
            \multirow{8}{*}{Disease Detection} & RSNA & A public medical dataset  includes chest X-rays from 1,218 pneumonia patients for object detection to identify and localize pneumonia.   & 0 / 32,024 / 0 \\
            & SLAKE &  A public medical dataset includes comprehensive semantic labels and bounding boxes annotated by experienced physicians, making it suitable for object detection tasks involving various human body parts.   & 580 / 0 / 0 \\ 
            & M3D-Seg  &  A public medical dataset includes image-mask-text triplets and 3D box coordinates derived from mask annotations, supporting object detection tasks with detailed region labeling.  & 122,644 / 799 / 0 \\
            \midrule
            \multirow{3}{*}{Landmark Detection} & IEEE ISBI Challenge 2015 Dataset & A public dataset of 400 cephalometric radiograph samples, widely used for cephalometric landmark detection, includes 19 landmarks annotated as 2D coordinates by two doctors.   & 2,850 / 0 / 4,750 \\
            
            \bottomrule 
        \end{tabular}
    }
\end{table*}

\subsubsection{Data Engine for Feature Alignment}
During the feature alignment stage, we use approximately 3 million image-text pairs, including 2.5 million from English medical vision-language datasets and 0.5 million from Chinese medical vision-language datasets. The details of acquiring these datasets are as follows.

For English medical vision-language datasets, 1.6 million image-text pairs are obtained by filtering and selecting from PMC-OA~\cite{lin2023pmc}, which contains paired images and captions collected from PubMed Central. Additionally, due to the limited quantity and variety of medical image datasets, we generate some synthetic training data, containing 0.9 million question-answer pairs. Specifically, we select pairs of different images and text as a group, using GPT-3.5 Turbo to answer their similarities and differences based on their corresponding titles and descriptions. Then, the selected images and corresponding generated answer consist a training sample. These training samples help the model better learn domain-specific knowledge, such as specialized terminology and concepts in the medical field.

For Chinese medical vision-language datasets, we select 0.3 million image-text pairs from ChiMed-VL-Alignment~\cite{liu2023qilin}, where each pair includes contextual information or a description of the image. Then, similar to the English vision-language datasets, we generate 0.2 million training samples with the same operation.

\subsection{Instruction-Tuning Stage}
Multi-task learning is crucial for achieving general artificial intelligence. In the instruction-tuning stage, we fine-tune UMIT for improving performance on multiple medical tasks. We select five downstream tasks, each of which has significant practical application value. Medical image report generation provides radiologists with high-quality draft reports, thereby significantly improving diagnostic efficiency~\cite{liu2024factual}. Medical VQA not only enhances diagnostic accuracy for clinicians but also improves patients’ understanding of medical information and advances medical education and research~\cite{zhang2023pmc}. Medical image classification plays a crucial role in disease diagnosis and lesion detection~\cite{chong2023category}. Disease detection, driven by text prompts, allows for the localization of target objects within images, identifying specific lesion areas~\cite{qin2022medical}. Landmark detection automatically identifies predefined anatomical landmarks within medical images, providing robust support for precise medical image analysis~\cite{chen2021deep}. During this stage, We freeze the pre-trained vision encoder and fine-tune the linear layer and language model. Thus, UMIT can better understand medical instructions and deal with medical tasks.

\subsubsection{Data Engine for Instruction-Tuning}
We select 18 publicly available datasets from previous work for training and test (details of these datasets are shown in Table~\ref{tab:Fine-tuning}). To help the model better understand task requirements and make necessary adjustments when handling different tasks (e.g., different output formats), we also design multiple instruction templates. The details are as follows.

\noindent\textbf{Medical VQA.} For the medical VQA task, we use a total of five datasets, including VQA-RAD~\cite{lau2018dataset}, SLAKE~\cite{liu2021slake}, PathVQA~\cite{he2021towards}, ChiMedVL-Instruction~\cite{liu2023qilin} for 2D images, and M3D-VQA~\cite{bai2024m3d} for 3D images. These datasets cover radiological data from various anatomical structures as well as pathological data capturing details of the body and specific tissues. In addition, some of these datasets include both open-ended and close-ended data. For open-ended data, we treat it as a generation task. Since the answers in VQA datasets are usually short, we design the following instruction template:  
\begin{tcolorbox}[colframe=black, colback=whitesmoke, coltitle=black, boxrule=0.3mm, boxsep=0.5mm,before skip=4pt, after skip=4pt]
\textless image\textgreater given the image, please provide a brief answer to \textless question\textgreater
\end{tcolorbox}
However, closed-ended data defines a set of predefined answer options, thus we treat it as a classification task, providing the candidate answers in the instruction: 
\begin{tcolorbox}[colframe=black, colback=whitesmoke, coltitle=black, boxrule=0.3mm, boxsep=0.5mm,before skip=4pt, after skip=6pt]
\textless image\textgreater given the image, choose one option from the \textless option1\textgreater \textbackslash n \textless option2\textgreater  to answer:\textless question\textgreater
\end{tcolorbox}

\noindent\textbf{Classification.} In the classification task, we collect seven datasets from MedMNIST~\cite{yang2023medmnist}, including colon pathology with nine categories, chest X-Ray for pneumonia detection (binary classification), dermoscopy images with seven typical pigmented skin lesions, retinal OCT images with four retinal diseases, breast ultrasound for detecting the presence of diseases, blood cell microscopy images with eight types of normal cells, and abdominal CT images with 11 organs. For the instruction of classification task, we adopt the same prompt template as used in VQA close-ended tasks.

\noindent\textbf{Report Generation.}  For medical report generation, we collect chest X-ray images IU-Xray~\cite{demner2016preparing}, CT images from M3D-Cap~\cite{bai2024m3d}, and clinical photos from PEIR GROSS~\cite{pavlopoulos2019survey}. The medical report generation task is similar to the image captioning task as both of them involve generating corresponding text based on an image. However, medical report generation requires more specialized knowledge compared to traditional image captioning, and the generated text is also more complex. Therefore, when designing instructions to train the model, we require the model to focus on identifying abnormalities in the report. Thus, we design the following instruction template:
\begin{tcolorbox}[colframe=black, colback=whitesmoke, coltitle=black, boxrule=0.3mm, boxsep=0.5mm,before skip=4pt, after skip=6pt]
\textless image\textgreater given the image, please review the image and create a report that assesses any abnormalities.
\end{tcolorbox}

\noindent\textbf{Disease Detection.} We use RSNA~\cite{gabruseva2020deep} and SLAKE~\cite{liu2021slake} for 2D image-based disease detection, and follow the template for object detection used in the Qwen2-VL:
\begin{tcolorbox}[colframe=black, colback=whitesmoke, coltitle=black, boxrule=0.3mm, boxsep=0.5mm,before skip=4pt, after skip=4pt]
Find \textless object\_ref\_start\textgreater \textless object\textgreater \textless object\_ref\_end\textgreater in this image.
\end{tcolorbox}
\noindent The label is required to follow this template: \textit{\textless box\_start\textgreater (x\_1,y\_1), (x\_2,y\_2) \textless box\_end\textgreater}.

For 3D image-based disease detection, we use M3D-Seg~\cite{bai2024m3d}. Due to the additional values in the $Z$ axis, we design the following instruction template: 
\begin{tcolorbox}[colframe=black, colback=whitesmoke, coltitle=black, boxrule=0.3mm, boxsep=0.5mm,before skip=4pt, after skip=4pt]
Find the \textless object\textgreater, please respond with a 3D bounding box.
\end{tcolorbox}
\noindent Similarly, the response is required to follow the format:
\textit{[(x\_1,y\_1,z\_1),(x\_2,y\_2,z\_2)]}

\begin{table*}[t]
\raggedright
    \centering
    \caption{Evaluation on 2D image-based medical VQA.}
    \label{vqa_en}
    \scalebox{1.0}{
        \begin{tabular}{cccccccccc} %
            \toprule 
            \multirow{2}{*}{\textbf{Model}} & \multicolumn{3}{c}{\textbf{SLAKE} (en)} & \multicolumn{3}{c}{\textbf{VQA-RAD}} & \multicolumn{3}{c}{\textbf{Path-VQA}} \\ 
             &  open & close & total & open & close & total & open & close & total \\
             \midrule
             Qwen2-vl & 3.3 & 41.45  & 19.28 & 4.35 & 43.75 & 21.44 & - & - & 2.15 \\
             \midrule
             BiomedCLIP & 84.3 & 88.9 & 86.1 & 67.0 & 78.5 & 72.7 &- & -& -\\
             PubMedCLIP & 78.4 & 82.5 & 80.1 & 60.1 & 80.0 &72.1 &- & -& - \\
             \midrule
             LLaVA-Med(LLaVA) & 83.1 & 85.3  & 84.0 & 61.5 & 84.2 & 75.2 & 38.0 & 91.2 & 64.7 \\
             LLaVA-Med(Vicuna) & 84.7 & 83.2  & 84.1 & 64.4 & 82.0 & 75.0  & 38.9 & 91.7 & 65.3  \\
             LLaVA-Med(Bio-CLIP)  & \textbf{87.1} & 86.8  & \underline{87.0} & 64.8 & 83.1 & 75.8 & 39.6 & 91.1 & 65.4  \\
             PeFoMed & 62.6 & 87.1  & 77.4 & \underline{77.8} & 88.7 & 82.1 & 35.7 & 91.3 & \underline{68.6} \\
             BiomedGPT-S & 66.5 & 73.3  & - & 57.8 & 13.4 & - & 10.7 & 84.2 & - \\
             BiomedGPT-M & 78.3 & 86.8  & - & 53.6 & 79.8 & - & 12.5 & 85.7 & - \\
             BiomedGPT-B & 84.3 & 89.9  & - & 60.9 & 81.3 & - & 28.0 & 88.0 & - \\
             \midrule
             UMIT-B & 81.39 & \underline{93.57}  & 86.71 & 72.5 & \underline{93.53} & \underline{82.29} & - & - & 63.71  \\
             UMIT & \underline{84.47} & \textbf{93.93}  & \textbf{89.2} & \textbf{79.87} & \textbf{94.38} & \textbf{86.62} & - & - & \textbf{69.29}\\
            \bottomrule 
        \end{tabular}
    }
\end{table*}

\noindent\textbf{Landmark Detection.} For landmark detection, we use the publicly available IEEE ISBI Challenge 2015 Dataset~\cite{lindner2016fully} for training and test. To enhance the flexibility of landmark detection, we specify the landmarks to be detected through textual descriptions. Compared to traditional methods, this approach can adapt to different requirements in practical applications and reduce computational costs. We design the following instruction template: 
\begin{tcolorbox}[colframe=black, colback=whitesmoke, coltitle=black, boxrule=0.3mm, boxsep=0.5mm,before skip=4pt, after skip=1pt]
\textless image\textgreater given the image, find the \textless landmark\_name\textgreater, the response is given in the format of [x,y].
\end{tcolorbox}

\section{Experiment}

\subsection{Implementation Details}
The model is implemented in PyTorch, and training occurs in parallel across 8 NVIDIA A100 GPUs with 80 GB memory. All our models are trained using the AdamW optimizer and leverage ZeRO-2 optimization enabled by DeepSpeed to optimize memory usage and accelerate the training process. In the first stage, we train UMIT for one epoch. In the second stage, we first train UMIT on all data for four epochs to obtain the base model (referred to as UMIT-B in the following section). Then, to align with previous models, such as BiomedGPT~\cite{zhang2024generalist}, we fine-tune UMIT on each task for the same number of epochs.

\begin{table}[t]
\raggedright
    \centering
    \caption{Evaluation on Chinese-based and 3D image-based medical VQA.}
    \label{vqa2}
    \scalebox{1.0}{
        \begin{tabular}{ccccc} %
            \toprule 
            \multirow{2}{*}{\textbf{Model}} & \multicolumn{2}{c}{\textbf{SLAKE} (zh)} & \textbf{CM-VL-I} & \textbf{M3D-VQA} \\ 
             &  open & close & open & open \\
             \midrule
             Qwen2-vl & 11.2 & 25.27  & 8.13 &-  \\
             \midrule
             RadFM & - & - & - & 17.79  \\
             M3D & - & - & - & \underline{75.78}  \\
             \midrule
             UMIT & \textbf{77.36} & \textbf{93.69} & \textbf{56.35} & \textbf{77.55} \\
            \bottomrule 
        \end{tabular}
    }
\end{table}

\begin{table*}[t]
\raggedright
    \centering
    \caption{Evaluation on classification.}
    \label{classi}
    \scalebox{0.85}{
        \begin{tabular}{cccccccc} %
            \toprule 
            \textbf{Model} & \textbf{PathMNIST} & \textbf{DermaMNIST} & \textbf{OCTMNIST} & \textbf{PneumoniaMNIST} & \textbf{BreastMNIST} & \textbf{BloodMNIST} & \textbf{OrganCMNIST} \\ 
            \midrule 
             Qwen2-vl & 15.6 & 11.3  & 12.1 & 62.5 & 30.77 & 9.07 & 21.31  \\
             \midrule
             Specialist SOTA & \underline{94.2} & \underline{79.8} & 78.2 & \underline{96.1} & \textbf{91.3} & 96.6 & \underline{92.2}\\
             \midrule
             BiomedGPT-S & 89.4 & 75.2  & 79.5 & 91.8 & 84.6 & 94.2 & 92.2 \\
             BiomedGPT-M & 92.1 & 78.0  & \underline{81.9} & 93.4 & 87.8 & 97.2 & 92.3  \\
             BiomedGPT-B & 92.6 & 78.6 & 81.6 & \textbf{96.7} & \underline{87.8} & \underline{97.7} & \textbf{93.1} \\
             \midrule
             UMIT-B & 85.65 & 72.37 & 62.6 & 84.46 & 83.97 & 86.64 & 83.05 \\
             UMIT & \textbf{95.4} & \textbf{82.54} & \textbf{82.9} & 90.62 & 87.36 & \textbf{98.46} & 90.77\\
            \bottomrule 
        \end{tabular}
    }
\end{table*}

\subsection{Results on Medical VQA}
In the 2D image-based medical VQA task, we evaluate the proposed UMIT on three English-only datasets. We use Qwen2-VL as the baseline and compare our model with previous CLIP-base and LLM-based SOTA methods in medical imaging. Specifically, we select two CLIP-based methods (\textit{i.e.}, BiomedCLIP and PubMedCLIP~\cite{eslami2023pubmedclip}) and three LLM-based models (\textit{i.e.}, LLaVA-Med~\cite{li2024llava}, PeFoMed~\cite{he2024pefomed}, and BiomedGPT). We use accuracy as the primary evaluation metric. As shown in Table~\ref{vqa_en}, our model significantly outperforms the baseline model Qwen2-VL. Two possible reasons may explain the poor performance of Qwen2-VL: 1) its training datasets contain very few medical-related examples; 2) it tends to generate longer answers than ground truth. Although the architecture of UMIT is based on Qwen2-VL, UMIT-B achieves a significant performance improvement, which further highlights the effectiveness of proposed training strategy and instruction templates. Furthermore, we compare the results of UMIT-B with other models. As shown in the results for the SLAKE and VQA-RAD datasets, UMIT-B achieves the best performance on closed-ended questions. Although UMIT-B does not reach the state-of-the-art for open-ended questions, its performance remains competitive. For the Path-VQA dataset, since the official dataset does not provide a split between closed-ended and open-ended questions, we treat all data as open-ended. Even so, it’s clear that our model shows considerable potential while other methods perform poorly on open-ended questions. Additionally, as shown in the last row of Table~\ref{vqa_en}, UMIT achieves optimal performance across all datasets, except for the open-ended questions on SLAKE. However, it should be noted that LLaVA-Med measures accuracy differently for open-ended and closed-ended questions: they use token recall rate for open-ended questions and conventional classification accuracy for closed-ended questions. 



As shown in Table~\ref{vqa2}, we also conduct experiments in other three datasets: two Chinese 2D image datasets SLAKE and CM-VL-I, and one 3D image dataset M3D-VQA. Since there have been no previous studies specifically focused on the two adopted Chinese 2D image datasets, we only compare UMIT with the baseline method Qwen2-VL. The results in Table~\ref{vqa2} show that UMIT also delivers impressive results in Chinese data. However, the performance on CM-VL-I is relatively poor, likely due to the complexity of this dataset. The CM-VL-I dataset encompasses both medical images and medical scene images and the answers are verbose, which may contribute to the challenges in achieving high performance. For the 3D image-based dataset, we compare our model with previous SOTA VLMs, including RadFM~\cite{wu2023towards} and M3D~\cite{bai2024m3d}. RadFM supports both 2D and 3D image analysis, while M3D is specialized for 3D image analysis. The results show that UMIT still achieves better performance, demonstrating that multi-task learning allows it to share knowledge across tasks.


\subsection{Results on Classification}
We evaluate UMIT on seven classification datasets and make comparison with both general VLMs and domain-specific models. The experimental results are shown in Table~\ref{classi}. Similar to the Medical VQA task, we use accuracy as the evaluation metric. Compared to the baseline model Qwen2-VL, UMIT shows significant performance improvement, which highlights its stronger domain adaptability. Clearly, UMIT achieves optimal performance on most datasets, even surpassing some specialized models. As in previous work, we use $28\times 28$ images as input for UMIT-B. However, this setup clearly had a negative impact on our model. In the proposed method, there is only one token left for the input image through the vision encoder, which may limit its performance. Although UMIT-B with $28\times 28$ input size achieves the subpar performance, they are still quite impressive compared with previous methods.


\subsection{Results on Report Generation}

As shown in Table~\ref{cap_2d} and Table~\ref{cap_3d}, we conduct experiments on 2D image datasets IU-Xray and PEIR GROSS and 3D image dataset M3D-CAP for the report generation task. We evaluate our method with commonly-used metrics: METEOR, ROUGE-L, CIDEr, and BLEU.

When evaluating on 2D image datasets, we select both SOTA general VLMs and domain-specific models as compared methods. The results presented in Table~\ref{cap_2d} show that UMIT outperforms previous LLM-based models on both IU-Xray and PEIR GROSS in METEOR and ROUGE-L metrics. This indicates that our model has clear advantages in content coherence and semantic similarity. We observe that the CIDEr score of UMIT is slightly lower since CIDEr emphasizes diversity and novelty, which is less relevant to medical report generation. Therefore, we believe this will not have a significant impact on the overall performance of our model. Moreover, our approach outperforms specialized models on the PEIR GROSS dataset and performs similarly on the IU-Xray dataset. This suggests that our model effectively utilizes multi-domain knowledge and contextual information, improving accuracy. 

In the evaluation of 3D image-based medical report generation as presented in Table \ref{cap_3d}, UMIT shows competitive performance compared to previous models. On M3D-Cap, UMIT achieves a BLEU score of 15.63, which is the highest among all the evaluated models, indicating its superiority in generating text that closely matches the reference reports on a lexical level. However, in ROUGE-L and METEOR metrics, UMIT scores 18.5 and 12.22 respectively, placing it slightly below the top performer M3D(mlp). Despite not leading in all metrics, UMIT's highest BLEU score signifies its strength in accurately reproducing key terms and phrases pertinent to medical reports from 3D images, which is crucial for effective communication in medical contexts.

\begin{table*}[t]
\raggedright
    \centering
    \caption{Evaluation on 2D image-based medical report generation.}
    \label{cap_2d}
    \scalebox{1.0}{
        \begin{tabular}{ccccccc} %
            \toprule 
             \multirow{2}{*}{\textbf{Model}} & \multicolumn{3}{c}{\textbf{IU-Xray}} & \multicolumn{3}{c}{\textbf{PEIR GROSS}} \\ 
            & ROUGE-L & METEOR & CIDEr & ROUGE-L & METEOR & CIDEr \\
            \midrule
             Qwen2-vl & 6.9 & 1.6  & 5.3 & 6.7 & 2.1 & 4.1  \\
             \midrule
             Specialist SOTA & 37.6 & 18.7 & 35.1 & 27.9 & 14.9 & 32.9 \\
             \midrule
             PeFoMed & 28.6 & \underline{15.7} & \textbf{46.2} & - & - & -\\
             BiomedGPT-S & 26.8 & 11.0  & 29.6 & 25.8 & 12.0 & 22.0  \\
             BiomedGPT-M & 28.0 & 11.0 & 31.3 & 24.0 &14.7 & 25.8 \\
             BiomedGPT-B & 28.5 & 11.0 & \underline{40.1} & 36.0 & 15.4 & \textbf{122.7} \\
             \midrule
             UMIT-B & \underline{29.6} & 13.9 & 31.0 & \underline{38.4} &\underline{19.5} & 67.6 \\
             UMIT & \textbf{30.3} & \textbf{16.0} & 34.3 & \textbf{42.6} &\textbf{22.9} & \underline{107} \\
            \bottomrule 
        \end{tabular}
    }
\end{table*}

\begin{table}[t]
\raggedright
    \centering
    \caption{Evaluation on 3D image-based medical report generation.}
    \label{cap_3d}
    \scalebox{1.0}{
        \begin{tabular}{cccc} %
            \toprule 
            \multirow{2}{*}{\textbf{Model}} & \multicolumn{3}{c}{\textbf{M3D-Cap}} \\ 
             &  BLUE & ROUGE-L & METEOR \\
             \midrule
             RadFM & 12.23 & 16.49 & 11.57   \\
             M3D(linear) & 14.49 & 19.25 & 14.11   \\
             M3D(mlp) & \underline{15.15} & \textbf{19.55} & \textbf{14.38}   \\
             \midrule
             UMIT & \textbf{15.63} & \underline{18.5} & \underline{12.22}  \\
            \bottomrule 
        \end{tabular}
    }
\end{table}

\subsection{Results on Disease Detection}
On 2D disease detection dataset RSNA, UMIT demonstrates a notable performance with an IoU score of 0.22, marking it as the second-best model. This result represents a significant improvement compared to general vision-language models Qwen2-VL and MiniGPT-v2, which achieve IoU scores of 0.10 and 0.13, respectively. Notably, MiniGPT-Med, optimized specifically for radiology image analysis, achieves the best performance with an IoU score of 0.26. Despite this, UMIT's results highlight its effectiveness in adapting to disease detection.

When evaluated on the 3D image-based disease detection dataset M3D-Seg, UMIT further asserts its superiority with a remarkable score of 67.21. UMIT outperforms both variants of M3D: one with a frozen vision encoder (M3D-F) scoring 30.05 and the other with a unfrozen vision encoder (M3D-U) achieving 49.66. The advantage of UMIT in 3D image-based disease detection is even more pronounced, showing an approximate improvement of 17.5\% over the best-performing M3D configuration. This indicates that UMIT not only excels in adapting to 2D medical images but also significantly advances the state-of-the-art in handling complex 3D medical imaging data, demonstrating its robustness and versatility across different medical imaging tasks.


\subsection{Results on Landmark Detection}
Since there is no LLM-based models for landmark detection, we only compare UMIT with the baseline model Qwen2-VL and the SOTA method FDGR-Net~\cite{li2024fdgr}. As shown in Table~\ref{landmark}, our results were slightly lower due to the limited dataset, with only 150 images per landmark for training. However, we still achieved significant improvements over the baseline. Although there were few images specifically for landmark detection in the training process, there were similar images used for other tasks. This could have allowed the model to learn relevant knowledge. This shows that the phase for multi-task instruction-tuning can effectively enhance model performance by leveraging related knowledge from different tasks.

\begin{table}[t]
\raggedright
    \centering
    \caption{Evaluation on 2D image-based disease detection.}
    \label{detection_2d}
    \scalebox{0.85}{
        \begin{tabular}{ccccc} %
            \toprule 
             \textbf{Dataset} & Qwen2-vl & MiniGPT-v2 & MiniGPT-Med & UMIT\\
             \midrule
             RSNA & 0.10 & 0.13 & \textbf{0.26}  & \underline{0.22} \\
             
            \bottomrule 
        \end{tabular}
    }
\end{table}

\begin{table}[t]
\raggedright
    \centering
    \caption{Evaluation on 3D image-based disease detection.}
    \label{detection_3d}
    \scalebox{0.9}{
        \begin{tabular}{cccc} %
            \toprule 
             \textbf{Dataset} & M3D-F & M3D-U & UMIT \\
             \midrule
             M3D-Seg  & 30.05 & 49.66 & \textbf{67.21}  \\
             
            \bottomrule 
        \end{tabular}
    }
\end{table}

\begin{table}[t]
\raggedright
    \centering
    \caption{Evaluation on landmark detection.}
    \label{landmark}
    \scalebox{1.0}{
        \begin{tabular}{cccccc} %
            \toprule 
             \multirow{2}{*}{\textbf{Model}} & \multirow{2}{*}{MRE} & \multicolumn{4}{c}{SDR} \\ 
              &  & 2mm & 2.5mm & 3mm & 4mm\\
             \midrule
             Qwen2-vl  & 10.15 & 12.6 & 21.91 & 33.1 & 39.84\\
             FDGR-Net & \textbf{1.45} & \textbf{75.16} & \textbf{82.53} & \textbf{88.56} & \textbf{94.79} \\
             \midrule
             UMIT-B & 2.07 & 57.73 & 71.84 & 81.63 & 92.37   \\
             UMIT & 1.97 & 63.79 & 76.95 & 85.32 & 93.21   \\
            \bottomrule 
        \end{tabular}
    }
\end{table}

\begin{figure*}[t]
    \centering
    \includegraphics[width=1.0\linewidth]{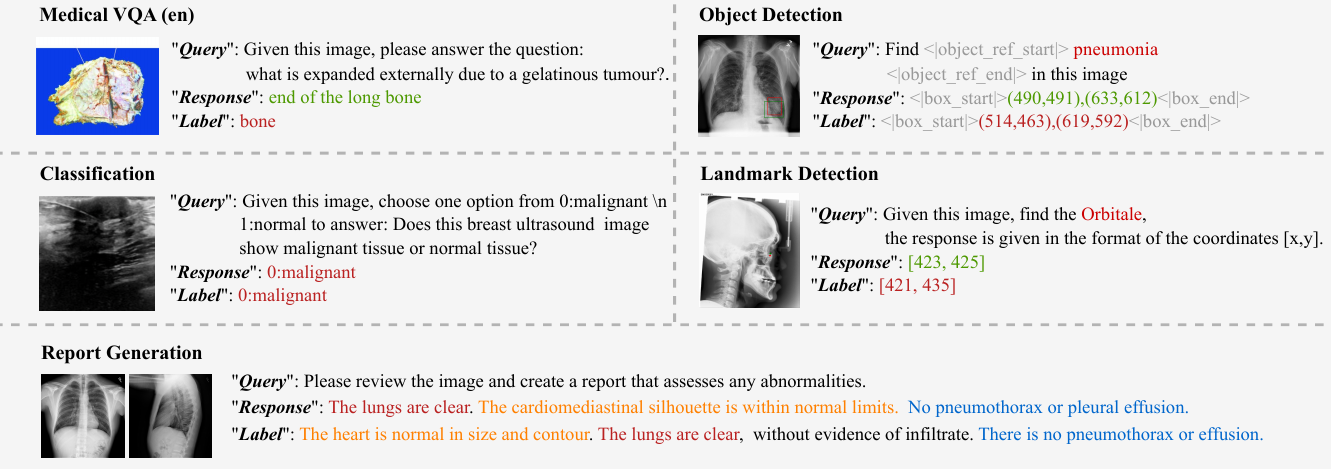}
\caption{Visualization of UMIT's results across five tasks.}
\label{visualize_re}
\end{figure*}

\subsection{Qualitative Evaluation}
As shown in the Figure~\ref{visualize_re}, we present partial visualization results across five tasks. 
In the classification tasks, including disease detection and tissue identification, the model consistently produces accurate results. 
For the VQA task, we present an example of an incorrect prediction. This highlights that, while the model performs well in general, it may still make errors when faced with more complex or ambiguous scenarios.
For landmark detection and disease detection tasks, although the predictions are not always perfectly accurate, the deviations are minimal. This indicates that the model has a strong understanding of the images and can recognize key features effectively. 
For the report generation task, we showcase a result based on multiple images. The model's predictions are highly coherent, and the conclusions align closely with the expected results. While the generated report lacks some of the depth found in the ground truth, it clearly demonstrates that the model can generate relevant and contextually sound conclusions. While slightly lacking in detail, the core output remains robust and reliable. More visualizations are shown in the supplementary materials.

\section{Conclusion}

In this paper, we introduce UMIT, a unified multi-modal, multi-task VLM designed to tackle various challenges in medical image analysis. Through an innovative two-stage training approach, we demonstrate that UMIT can effectively integrate diverse medical imaging modalities, while handling multiple tasks. Experimental results show that UMIT significantly outperforms state-of-the-art methods across multiple task, highlighting its strong generalizability. Overall, UMIT provides a powerful unified solution for medical imaging tasks.

{
    \small
    \bibliographystyle{ieeenat_fullname}
    \bibliography{main}
}

\clearpage
\setcounter{page}{1}
\maketitlesupplementary

\section{Qualitative Results}

\begin{figure}[t]
    \centering
    \includegraphics[width=1\linewidth]{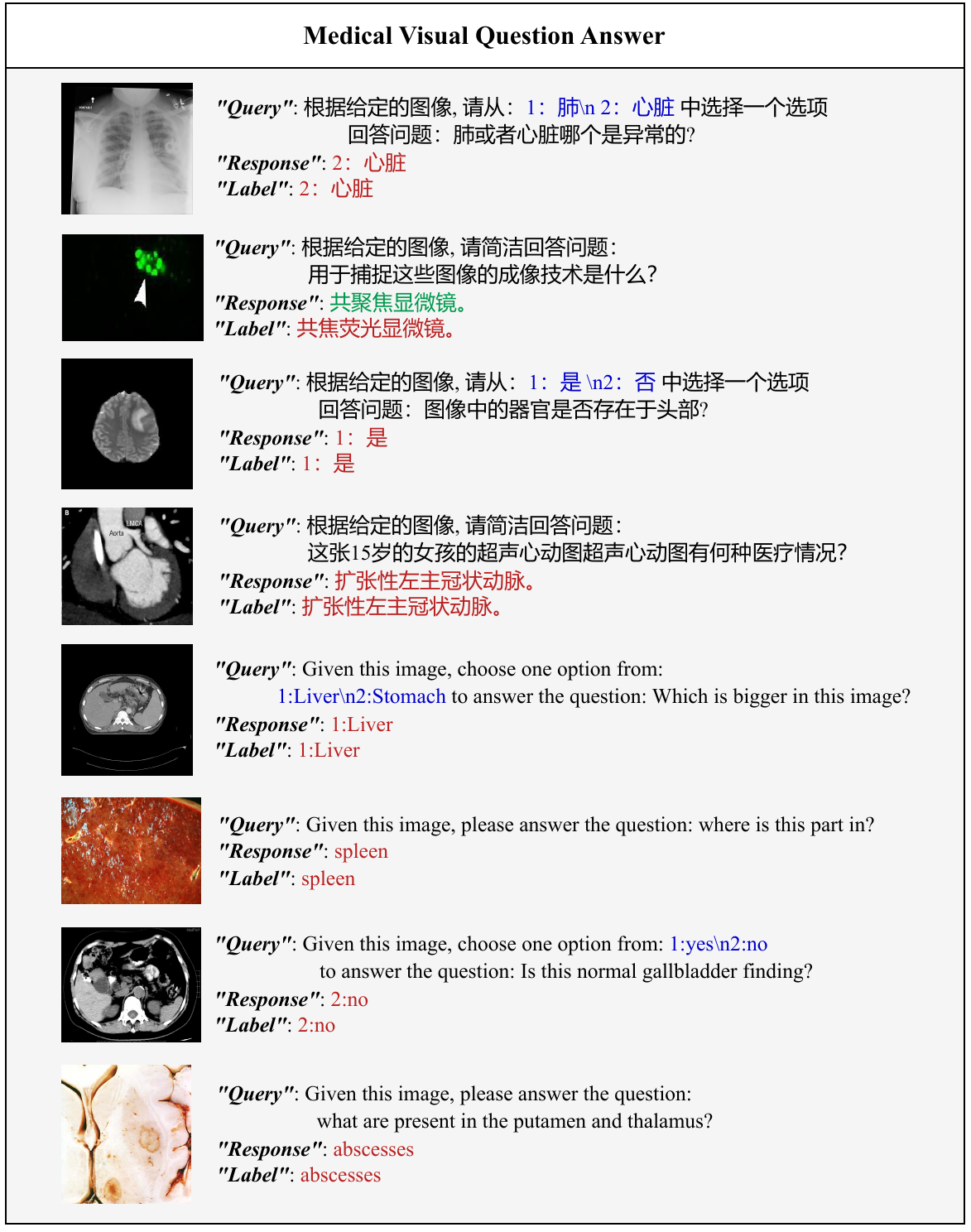}
\caption{Qualitative examples of Medical VQA task.}
\label{first}
\end{figure}

\begin{figure*}[htbp]
    \centering
    \includegraphics[height = 20.5cm,width=1\linewidth]{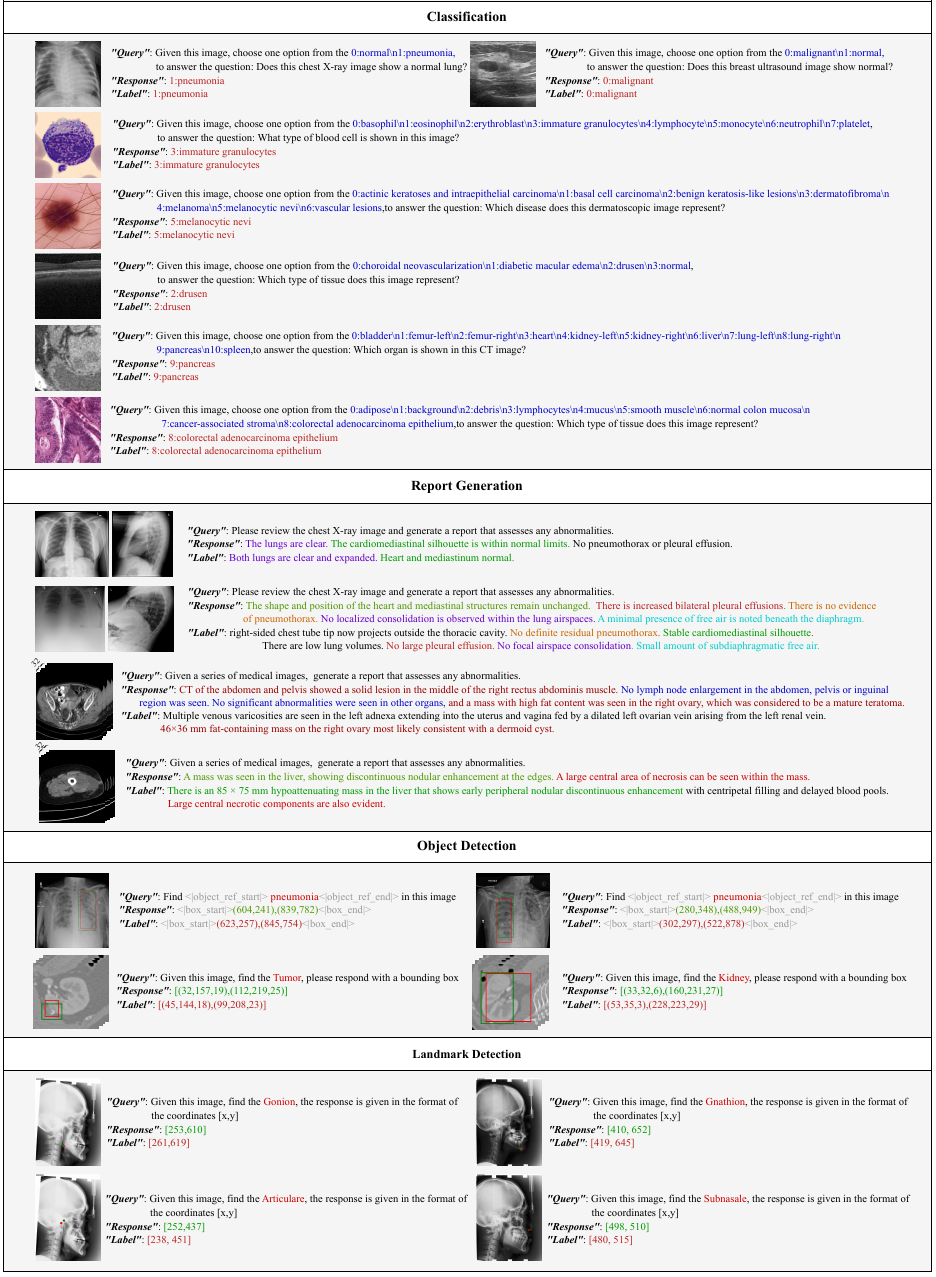}
\caption{Qualitative examples of classification task, report generation task, disease detection task and landmark detection task.}
\label{second}
\end{figure*}

In this section, we provide sufficient output results of UMIT to show its outstanding performance, as shown in Figure~\ref{first} and Figure~\ref{second}.

\noindent\textbf{Medical VQA.} As shown in Figure~\ref{first}, UMIT demonstrates its ability to understand questions and provide accurate answers in the Medical VQA task. It performs well across both Chinese and English questions, as well as for close-ended and open-ended formats. The model can effectively answer diagnostic questions (e.g., disease diagnosis) and factual questions (e.g., knowledge about imaging techniques). Notably, some predictions, while not identical to the ground truth labels, are still valid. For instance, in the second example, the label is 'confocal fluorescence microscope,' while the model predicts 'confocal microscope.' Since a confocal fluorescence microscope is a type of confocal microscope, the prediction, though less specific, is still correct. This indicates the model's strong generalization capability in understanding and answering questions.

\noindent\textbf{Classification.} The Figure~\ref{second} presents the visualization results of UMIT on seven classification task datasets. These datasets cover various types of tasks, including colorectal pathology, pneumonia detection, retinal diseases, and skin lesions. It can be observed that UMIT achieves accurate predictions in both binary and multi-class classification tasks.

\noindent\textbf{Report Generation.} Figure~\ref{second} qualitatively demonstrates the powerful report generation capability of UMIT. We use consistent color coding to highlight similar content between the predictions and the labels. It can be observed that UMIT accurately understands image content and diagnoses corresponding diseases, whether based on 2D or 3D images. While it occasionally generates additional content, this serves as further description of the images rather than introducing unnecessary or incorrect diagnoses.

\noindent\textbf{disease Detection.} Figure~\ref{second} presents the disease detection results of UMIT on 2D and 3D images. The 2D image dataset is primarily used for pneumonia detection, while the 3D dataset includes disease detection (e.g., locating tumors) and organ recognition (e.g., identifying the position of the kidneys). It can be observed that UMIT accurately detects the disease locations, further demonstrating the reliability and effectiveness of the model in disease detection tasks.

\noindent\textbf{Landmark Detection.} Figure~\ref{second} presents the visualization results of landmark detection based on landmark names. UMIT accurately identifies the corresponding locations in the image based on the given landmark names, demonstrating its strong cross-modal understanding of images and text, semantic comprehension of landmark names, and contextual reasoning capabilities. In medical scenarios, the model effectively integrates landmark names with image information to precisely detect anatomical structures or pathological features, further showcasing its generalization ability and practical value in multi-modal tasks.

\end{document}